\documentclass[prodmode,acmtist]{acmsmall} % Aptara syntax

\usepackage[ruled]{algorithm2e}
\renewcommand{\algorithmcfname}{ALGORITHM}
\SetAlFnt{\small}
\SetAlCapFnt{\small}
\SetAlCapNameFnt{\small}
\SetAlCapHSkip{0pt}
\IncMargin{-\parindent}

\usepackage{algorithmic}
\usepackage{amsmath}
\usepackage{graphics}
\usepackage{graphicx}  
\usepackage{subfloat}
\usepackage{subfig}
\usepackage{caption}
\usepackage{booktabs}
\usepackage{floatrow}
\newfloatcommand{capbtabbox}{table}[][\FBwidth]

\acmVolume{V}
\acmNumber{N}
\acmArticle{A}
\acmYear{YYYY}
\acmMonth{1}

\begin{document}

% Page heads
\markboth{N. Figueroa et al.}{A Combined Approach Towards Consistent Reconstructions.}

% Title portion
\title{A Combined Approach Towards Consistent Reconstructions of Indoor Spaces based on 6D RGB-D Odometry  and KinectFusion}
\author{Nadia Figueroa
\affil{Ecole Polytechnique Federale de Lausanne (EPFL)}
Haiwei Dong
\affil{University of Ottawa}
Abdulmotaleb El Saddik
\affil{University of Ottawa and New York University AD}}

\begin{abstract} 
We propose a 6D RGB-D odometry approach that finds the relative camera pose between consecutive RGB-D frames by keypoint extraction and feature matching both on the RGB and depth image planes. Furthermore, we feed the estimated pose to the highly accurate KinectFusion algorithm, which uses a fast ICP (Iterative-Closest-Point) to fine-tune the frame-to-frame relative pose and fuse the Depth data into a global implicit surface. We evaluate our method on a publicly available RGB-D SLAM benchmark dataset by Sturm et al. The experimental results show that our proposed reconstruction method solely based on visual odometry and KinectFusion outperforms the state-of-the-art RGB-D SLAM system accuracy. Moreover, our algorithm outputs a ready-to-use polygon mesh (highly suitable for creating 3D virtual worlds) without any post-processing steps. 
\end{abstract}

%\category{C.2.2}{Computer-Communication Networks}{Network Protocols}

\terms{Design, Algorithms, Performance}
\terms{3D Mapping, RGB-D Sensing, Visual Odometry}


\acmformat{Nadia Figueroa, Haiwei Dong and Abdulmotaleb El Saddik. 2013. A Combined Approach Towards Consistent Reconstructions of Indoor Spaces based on 6D RGB-D Odometry  and KinectFusion.}

\begin{bottomstuff}

Author's addresses: N. Figueroa is with the Learning Algorithms and Systems Laboratory, Ecole Polytechnique Federale de Lausanne (EPFL), Switzerland; H. Dong is with the School of Electrical Engineering and Computer Science, University of Ottawa, Ontario, Canada; A. El Saddik is with the School of Electrical Engineering and Computer Science, University of Ottawa and is visiting the Division of Engineering, New York University Abu Dhabi.
\end{bottomstuff}

\maketitle

\section{Introduction}
In this work, we aim at generating consistent reconstructions of indoor spaces using a freely moving hand held RGB-D sensor, that can lead to the analysis and interpretation of the 3D reconstructed indoor models by automatically generating smart representations, such as 3D CAD models (consequently 3D printed models) and 2D layouts. With the release of low-cost RGB-D sensors such as the Microsoft Kinect\footnote{Microsoft Kinect. http://www.xbox.com/en-us/kinect} and the PrimeSense 3D Sensor\footnote{Primesense 3D Sensor. http://www.primesense.com/}, the possibility of reconstructing our surroundings, objects and even ourselves in 3D has been reached. Thus, we present an algorithm that is targeted at using these devices. 

We propose a novel VO (Virtual Odometry) algorithm based on the RGB-D data of consecutive frames and \emph{KinFu Large Scale}, which shows promising results compared to a state-of-the-art RGB-D SLAM system. In contrast to the experiments presented by \cite{Kintinuous2}, our approach combines VO and ICP (Iterative-Closest-Point) as coarse-to-fine alignment method, i.e. VO provides a coarse alignment (rough rigid motion guess) between the consecutive camera poses and ICP fine-tunes (or corrects) this initial estimate. This procedure has been shown to generate consistent reconstructions from RGB-D data \cite{Figueroa-CRV2012}, as opposed to using switching strategies between different estimation approaches which might ignore good quality alignments due to the user-set threshold.

Our indoor 3D reconstruction algorithm is an iterative two-step procedure based on an adaptation of KinectFusion algorithm \cite{KinectFusion} to a novel 6D RGB-D odometry. We propose a robust 6D RGB-D odometry algorithm which considers useful information from both RGB and Depth images to estimate the 6D rigid motion of a moving RGB-D Sensor. The result of the estimated odometry is used as an initial transformation guess to the KinectFusion's fast ICP algorithm. As the original KinectFusion relies solely the ICP algorithm based on depth data, it is prone to camera pose estimation failures in the presence of planar surfaces with undescriptive 3D features. Thus, we present two contributions in this paper: (i) the novel 6D RGB-D odometry algorithm and (ii) an improvement to the KinectFusion algorithm by combining it with our proposed odometry estimation. Furthermore, we evaluate our proposed approach on a publicly available RGB-D benchmark dataset used for the evaluation of RGB-D SLAM systems \cite{Benchmark} and compare it with the publicly available results of the RGB-D SLAM presented by \cite{Endres}.

This paper is organized as follows. In Section \ref{sec:problem}, we present the problem formulation of the targeted application in this work and in Section \ref{sec:approach} we describe the proposed approach and present a detailed description of each component. In Section \ref{sec:evaluation}, we provide the evaluation results of our system applied on the RGB-D benchmark and present promising results. Finally, we conclude the whole paper in Section \ref{sec:conclusion}.

%%%%%%%%%%%%%%%%%%%%%%%%%%%%%%%%%%%%%%%%%%%%%%%%%%%%%%%%%%%%%%%%%%%%%%%%%%%%%%%%
\section{PROBLEM FORMULATION} 
\label{sec:problem}
The indoor mapping problem using a freely moving handheld sensing device is based on estimating the pose of the sensor (i.e. camera pose $C$) for each $k$-th frame from a recorded trajectory of the sensing device and simultaneously building a map of the environment with the estimated camera poses and acquired 3D representations of each frame. The 6DOF camera pose $C_k = (R_k,t_k)$ is a rigid body transformation matrix, where $R_k \in \mathbb{SO}_3$, $t_k \in \mathbb{R}^{3}$ and $k$ is the $k$-th camera pose for $k=1,...,N$ (N=total number of frames/poses). In order to generate the trajectory of successive camera poses, we need to estimate the rigid motion between consecutive camera poses $C_{k-1} \rightarrow C_{k}$.  This can be defined with a homogenous transformation $T_{k-1}^{k}=\left(R_{k-1}^{k},t_{k-1}^{k}\right)$ of a camera pose $C_{k-1}$ with respect to camera pose $C_k$ as $C_{k}=T_{k-1}^{k}\left(C_{k-1}\right)$.

By setting the initial camera pose $C_0$ to an identity matrix or a specified initial pose, any successive camera pose can be estimated by incrementally multiplying the rigid motion transformations between consecutive frames, thus $C_k$ can be re-written as
\begin{equation}
C_k = \prod_{i=1}^k T_{i-1}^{i} C_{0} = T_{k-1}^{k} \left(  \prod_{i=1}^{k-1} T_{i-1}^{i} C_{0}  \right).
\label{eq:incre}
\end{equation}

It can be inferred by Equation \ref{eq:incre} that if any estimated $T_{i-1}^{i}$ are erroneous or present drift, these errors are carried out throughout the complete trajectory. This error can be reduced and evenly distributed throughout the trajectory by identifying loop closures and optimizing the pose trajectories as in SLAM approaches. However, in this paper we propose an approach that estimates the incremental rigid motion $T_{i-1}^{i}$ with high accuracy using both RGB and Depth information in the pose estimation method and reaches the same accuracy as a SLAM-based approach. In the following section, we describe in detail our proposed approach. 

%%%%%%%%%%%%%%%%%%%%%%%%%%%%%%%%%%%%%%%%%%%%%%%%%%%%%%%%%%%%%%%%%%%%%%%%
\section{OUR PROPOSED INDOOR MAPPING ALGORITHM}
\label{sec:approach}
In order to obtain the best estimate of the incremental rigid motions $T_{k-1}^{k}$ between consecutive frames $C_{k-1} \rightarrow C_{k}$ we propose a two-step coarse-to-fine pose estimation approach. Initially, we introduce a novel visual odometry algorithm based on the RGB-D data of consecutive frames, which is thorougly described in Section \ref{sec:VO}. This resulting estimated pose from our VO is used as a coarse estimation, i.e. an initial guess to a fine estimation method (fast-ICP) used within the KinectFusion algorithm introduced by \cite{KinectFusion} (described in Section \ref{sec:kinfu}). 

\begin{figure}[h]
\vspace{-2mm}
\includegraphics[width=\linewidth]{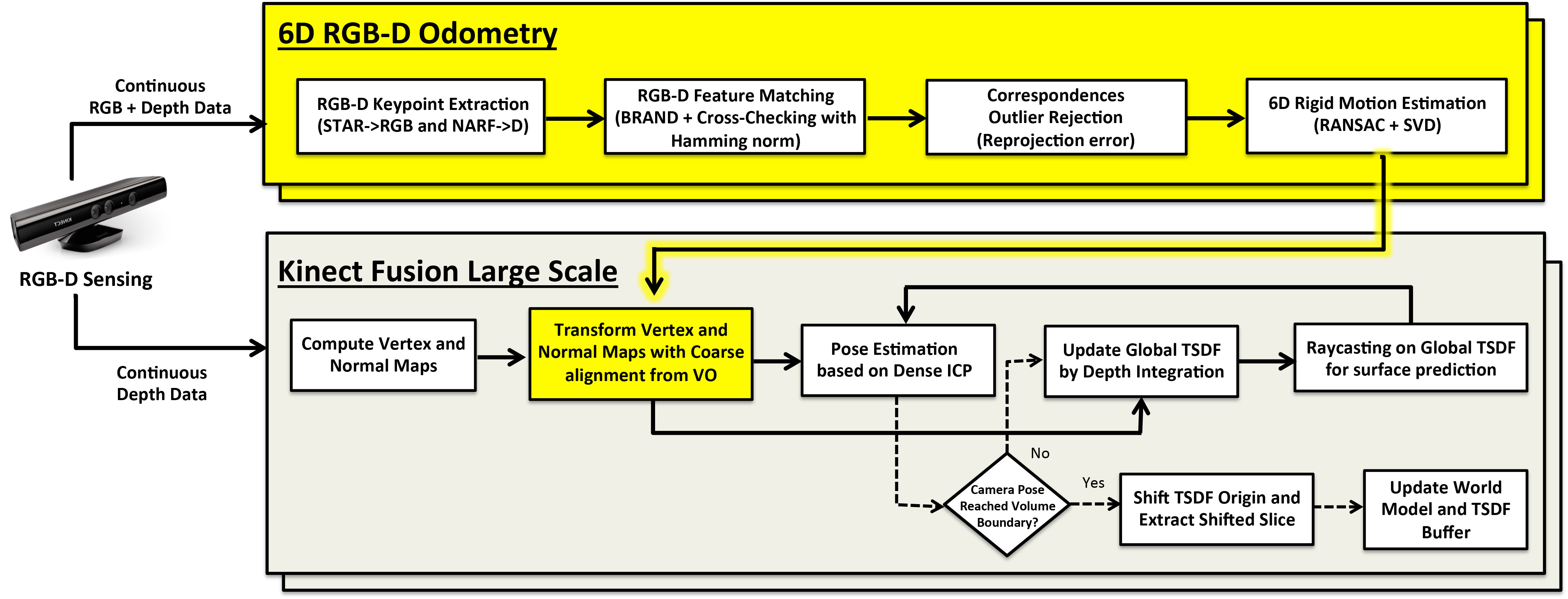}
\vspace{-2mm}
\caption{\small Schematic overview of our approach.}
\label{fig:schematic}
\end{figure}

The map is then incrementally updated in a global implicit model of the surface inferred from the aligned depth data and as the trajectory finalizes the model can be obtained as a 3D volumetric representation or a polygon mesh by applying the marching cubes algorithm on the 3D data. These steps are illustrated in a schematic overview of our approach in Fig. \ref{fig:schematic}.  The yellow blocks and yellow shadowed arrows represent our contributions. The dashed lines represent the KinectFusion extension for large scale environments implemented by Heredia and Favier$^{3}$.

\subsection{6D RGB-D Odometry}
\label{sec:VO}
Visual odometry (VO) is the process of estimating the incremental pose of a visual sensing device, either standalone or mounted on an agent (i.e. robot, human or vehicle), by using the captured images \cite{VO}. The standard VO approach is based on the following components: (i) feature detection, (ii) feature matching/tracking and (iii) outlier removal (generally based on RANSAC). In the past years, RGB images have been used to extract visual appearance information from the scenes with the well-known Scale Invariant Feature Transform (SIFT) \cite{Lowe-IJCV04} and Speed Up Robust Features (SURF) \cite{Bay-CVIU08} algorithms. However, when applying them on real world environments, several issues arise, such as variance in illumination, textureless objects, occlusion and surface reflectance which cause these descriptors to perform poorly. However, with the advent of RGB-D data we can combine the visual appearance and shape information in a fully descriptive and discriminative manner. Thus, in our approach we follow the standard VO workflow, with the novelty of applying the feature detection and feature tracking components on both RGB and Depth images (i.e. we use both visual and shape keypoint extraction algorithms and a combined RGB-D feature detector) and the 6DOF rigid motion is estimated using the 3D points corresponding to the extracted feature points. 

\subsubsection{RGB-D Keypoint Extraction}
In order to extract the most salient keypoints both on the RGB ($i_k \in I$ for $k=1..N$)  and Depth image ($d_k \in D$ for $k=1..N$) we compute a set of keypoints $k_{rgb}\in K_{rgb}$ from $i_k$ using the Center Surrounded Extrema keypoint detector introduced by \cite{STAR} and implemented as the STAR detector in the Open Source Computer Vision (OpenCV) library \cite{opencv_library} and another set of key points $k_{d}\in K_{d}$ from $d_k$ using the NARF (Normal Aligned Radial Feature) interest point detector which is applied by converting the depth date into a range image. Even though the NARF and STAR detectors identify different types of keypoints, this does not rule out the fact that a certain pixel region might be salient both with shape and texture, thus these two algorithms can provide the same key points. We define the final set of keypoints ($k \in K$) for an RGB-D frame ($i_k$ + $d_k$) is $ K = K_{rgb} \cup K_{d}$, where $\cup$ is the union of the two sets of keypoints. In our implementation, the union represents combining both sets and removing duplicate keypoints.

\subsubsection{RGB-D Feature Matching}
Recently, researchers have taken the task to correctly combine and fuse appearance and shape descriptions into a unique feature descriptor. The outcome has been the introduction of several feature vectors, namely the  Convolution k-means, EMK-Spin + EMK-SIFT, BRAND, CSHOT, PFHRGB, MeshHOG, VOSCH. 
A comparative overview of all of these previously mentioned feature vectors is presented in \cite{RAS-Ali}. We chose to use the novel BRAND (Binary Robust Appearance and Normals Descriptor) feature descriptor introduced by \cite{Brand}. It combines intensity and shape information in a binary bit string using a binary operator within a patch centered at a keypoint and has demonstrated to outperform the CSHOT feature descriptor, which is one of the dominant RGB-D feature descriptor algorithms. For a consecutive pair of RGB-D images $(i_{k-1}+d_{k-1}) \rightarrow (i_{k}+d_{k})$ we compute a set of BRAND feature descriptors $F_{k-1}$ and $F_{k}$ and apply a brute-force descriptor matcher with the Hamming norm and cross checking the correspondences, which results in a set of matches $m_i \in M_{k-1}^{k}$, where $i=1...N$ and $N$ is the total number of frame-to-frame matches. 

\subsubsection{Correspondences Outlier Rejection}
The corresponding frame-to-frame matches $M_{k-1}^{k}$ are still prone to wrong data associations, thus we further filter the matches using a RANSAC outlier rejection algorithm that finds the homography $H$ between the set of corresponding keypoints that compose $M_{k-1}^{k}$, transforms the key points of the $(k-1)$-th frame to the image plane of the $k$-th frame, computes the reproduction error between these sets of points and removes the correspondences which surpass a user defined reproduction threshold. The result is a set of minimal consistent corresponding matches $MM_{k-1}^{k}$ between $(i_{k-1}+d_{k-1}) \rightarrow (i_{k}+d_{k})$. Our acquired RGB-D images are already pre-calibrated, thus we can automatically compute $MM_{k-1}^{k}$ in the cartesian space; i.e. 3D points $P_{k-1}\rightarrow P_{k}$, where $p \in P$ and $p=(x,y,z)$. 

\subsubsection{6D Rigid Motion Estimation}
Once the final corresponding matches $MM_{k-1}^{k}$ are obtained, we compute the 6D rigid motion (inspired by RANSAC) by iteratively estimating the best transformation between the two sets of 3D points $P_{k-1}$ and $P_{k}$, corresponding to the remaining correspondences. We set a maximum number of hypotheses iterations $N$ where we randomly select $s$ samples from $MM_{k-1}^{k}$. We extract the 3D points of the $s$ corresponding matches and estimate a transformation hypothesis using the Umeyama Method \cite{Umeyama}.\footnote{Derivation provided in our online Appendix.} This yields to a transformation matrix $T_i=[R_i,t_i]$, where $i=1,...,N$. We then compute a transformation error $e_i$ corresponding to $T_i$ by computing the Root Mean Square (RMS) error based on the euclidean distances between the transformed set of points $P^{t}_{k-1} = T_{i}P_{k-1}$ and $P_{k}$. After computing $N$ transformation hypotheses, the transformation $T_i$ that produces the minimum error $e_i$ is chosen as the final rigid motion transformation hypothesis. The resulting rigid transformation ${T^{k}_{k-1}}_{rgbd}$ aligns $P_{k-1} \rightarrow P_{k}$ and consequently camera poses $C_{k-1} \rightarrow C_{k}$. The complete iterative rigid motion estimation approach is provided in the online Appendix.

\subsection{Camera Pose Fine Tuning and Surface Reconstruction with KinectFusion Large Scale}
\label{sec:kinfu}
As mentioned earlier, KinectFusion is based on incrementally fusing consecutive frames of depth data into a 3D volumetric representation of an implicit surface. This representation is the truncated signed distance function (TSDF) \cite{SDF}. The TSDF is basically a 3D point cloud stored in GPU memory using a 3D voxelized grid. The global TSDF is updated when a new depth image frame is acquired and the current camera pose is estimated. Initially, the depth image from the Kinect sensor is smoothed out using a bilateral filter \cite{eccv-12-qingxiong-yang}, which up-samples the raw data and fills the depth discontinuities. Then the camera pose of the current depth image frame is estimated with respect to the global model by applying a fast Iterative-Closest-Point (ICP) algorithm between the currently filtered depth image and a predicted surface model of the global TSDF extracted by ray casting. Once the camera pose is estimated, the current depth image is transformed into the coordinate system of the global TSDF and updated. Following we describe in detail the camera pose estimation method and the global TSDF update procedure.

\subsubsection{Camera Pose Estimation}
\begin{figure}[h]
        \centering
        \subfloat[freiburg1\_room sequence.]{\includegraphics[scale=0.12]{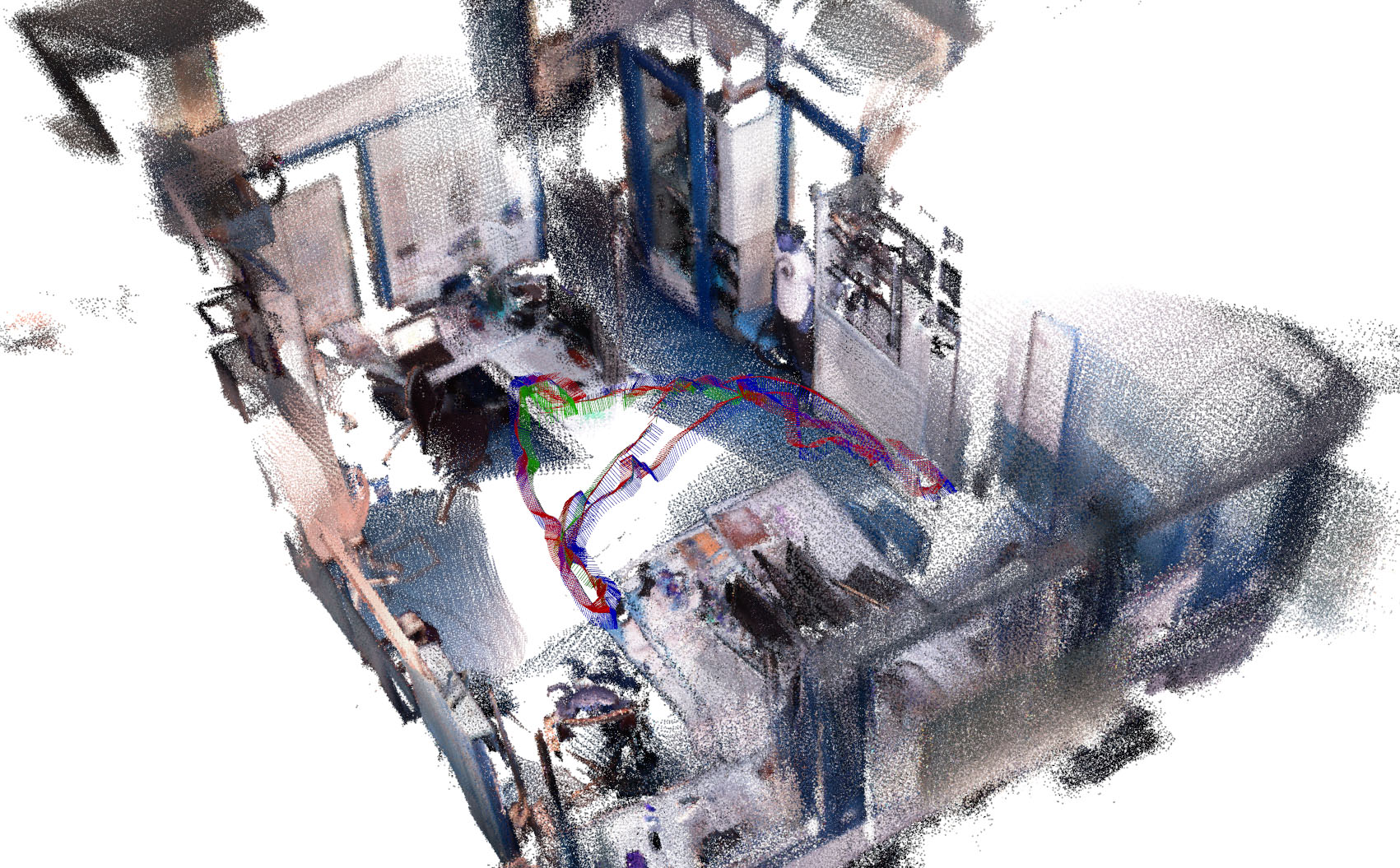}}  
        \subfloat[freiburg2\_desk sequence.]{ \includegraphics[scale=0.14]{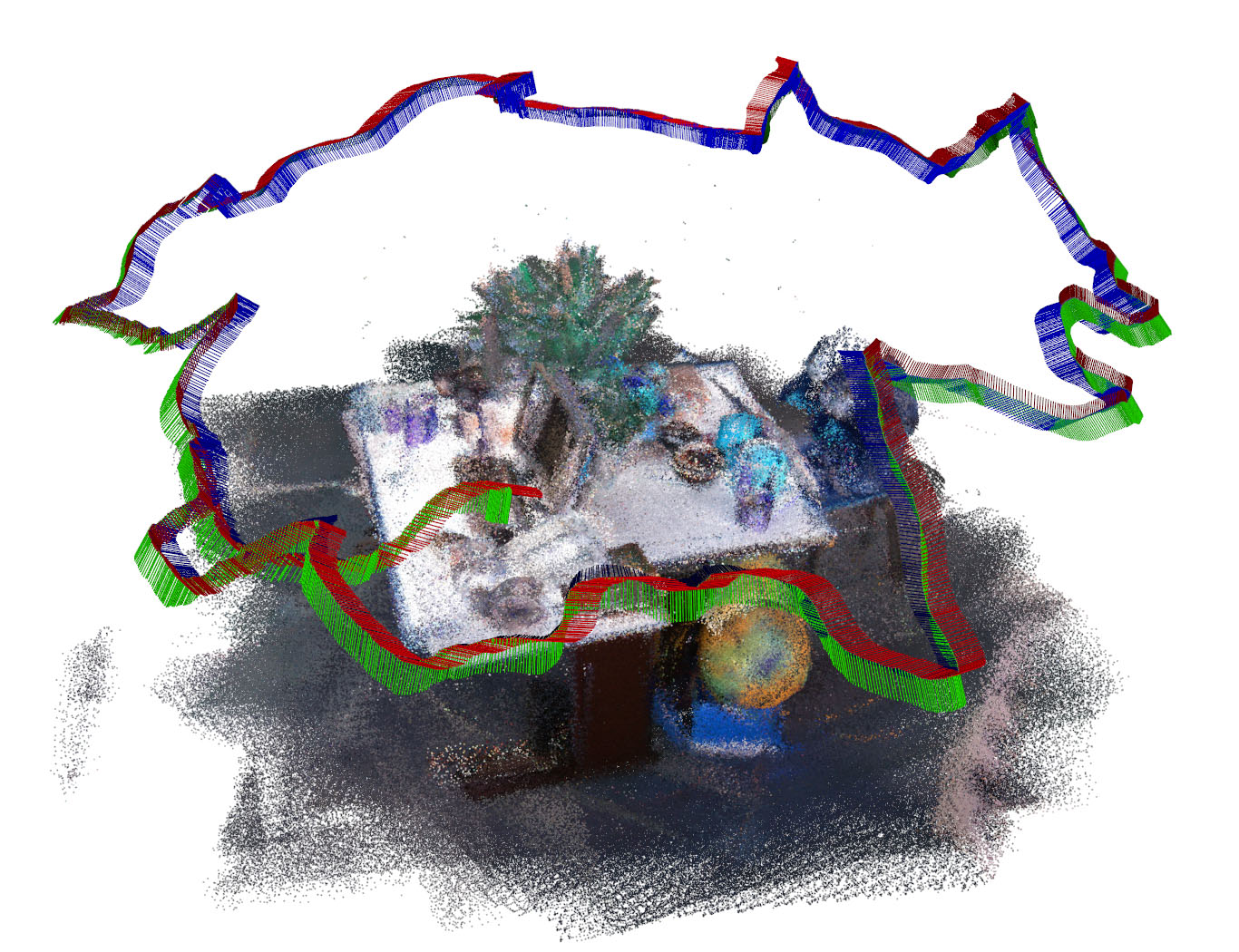}}
           \caption{\small Projected camera poses on a voxelized volumetric representation of the reconstructed map from the RGB-D Benchmark.}   
        \label{fig:trajectories}
\end{figure}

As mentioned earlier, we adapt the KinectFusion algorithm by initially transforming the new depth data with the coarse alignment estimated by 6D RGB-D odometry $T_{rgbd}$. Thus, we define the source points as $P_s = ({T^{k}_{k-1}}_{rgbd})^{-1}P_{k}$ and the target points $P_t$ as the predicted surface model of the global TSDF. Then we apply ICP to fine-tune the transformation between $P_s$ and $P_t$. The principal of the ICP algorithm is to find a data association between the subset of the source points ($P_{s}$) and the subset of the target points ($P_{t}$) \cite{Besl92}\cite{Chen91}. We use a special variant of the ICP-algorithm, the point-to-plane ICP  \cite{Zhengyou94}. It minimizes the error along the surface normal of the target points $n_t$, as in the following equation: 
\begin{equation}
T^{*}  =  \underset{T}{\mathrm{argmin}}\sum_{p_{s}\in P_{s}} \lVert n_t \cdot (T(p_{s})-p_{t})\rVert_{2}\\
\end{equation}
where $n_t \cdot (T(p_{s})-p_{t})$ is the projection of $(T(p_{s})-p_{t})$ onto the sub-space spanned by the surface normal ($n_t$). Now, for every $k$-th frame of the recorded trajectory, the incremental rigid motion is estimated as $T_{k-1}^{k}= (T^{*})^{-1}$. The resulting camera pose estimations of a freely moving handheld kinect recorded from the freiburg1\_room and freiburg2\_desk sequences can be seen in Fig. \ref{fig:trajectories}.

\subsubsection{Global TSDF Updating}
\label{sssec:global}
After computing transformation $T^{*}$, the new depth image is transformed into the coordinate system of the global TSDF by $T^{*}(P_s)$. The global model is represented in a voxelized 3D grid and integrated using a simple weighted running average. For each voxel, we have a value of signed distance for a specific voxel point $x$ as $d_{1}\left(x\right)$, $d_{2}\left(x\right)$, $\cdots$, $d_{n}\left(x\right)$ from $n$ depth images ($d_i \in D$) in a short time interval. To fuse them, we define n weights $w_{1}\left(x\right)$, $w_{2}\left(x\right)$, $\cdots$, $w_{n}\left(x\right)$. Thus, the weight corresponding point matching can be written in the form
\begin{equation}
w_{n}^{*}=\underset{k}{\mathrm{arg}}\sum_{k=1}^{n-1}\left\Vert W_{k}S_{k}-S_{n}\right\Vert _{2}
\end{equation}
where $S_{k+1}$ is the cumulative TSDF and $W_{k+1}$ is the weight functions after the integration of the current depth image frame, as described by \cite{SDF}. Furthermore, by truncating the updated weights to a certain value $W_{\alpha}$ a moving average reconstruction is obtained. 

\subsubsection{World Model Update and Volume Shifting}
In the original implementation of KinectFusion, the global TSDF is restricted to a fixed volume based on the resolution of the volumetric representation and the GPU memory capacity. In KinectFusion extension for large scale environments implemented by Heredia and Favier, where they allow the TSDF volume to move using a 3D cyclic buffer as the camera pose translation reaches a specified distance from the initial origin. Once the camera pose reaches this boundary, the origin of the TSDF volume is shifted and the slice of the 3D data that remains outside of the new volume is integrated to the world model. 

\subsection{Meshing with Marching Cubes}
\label{sssec:marchingcubes}
\begin{figure}
        \centering
        \subfloat[freiburg2\_desk sequence reconstruction.]{\includegraphics[scale=0.165]{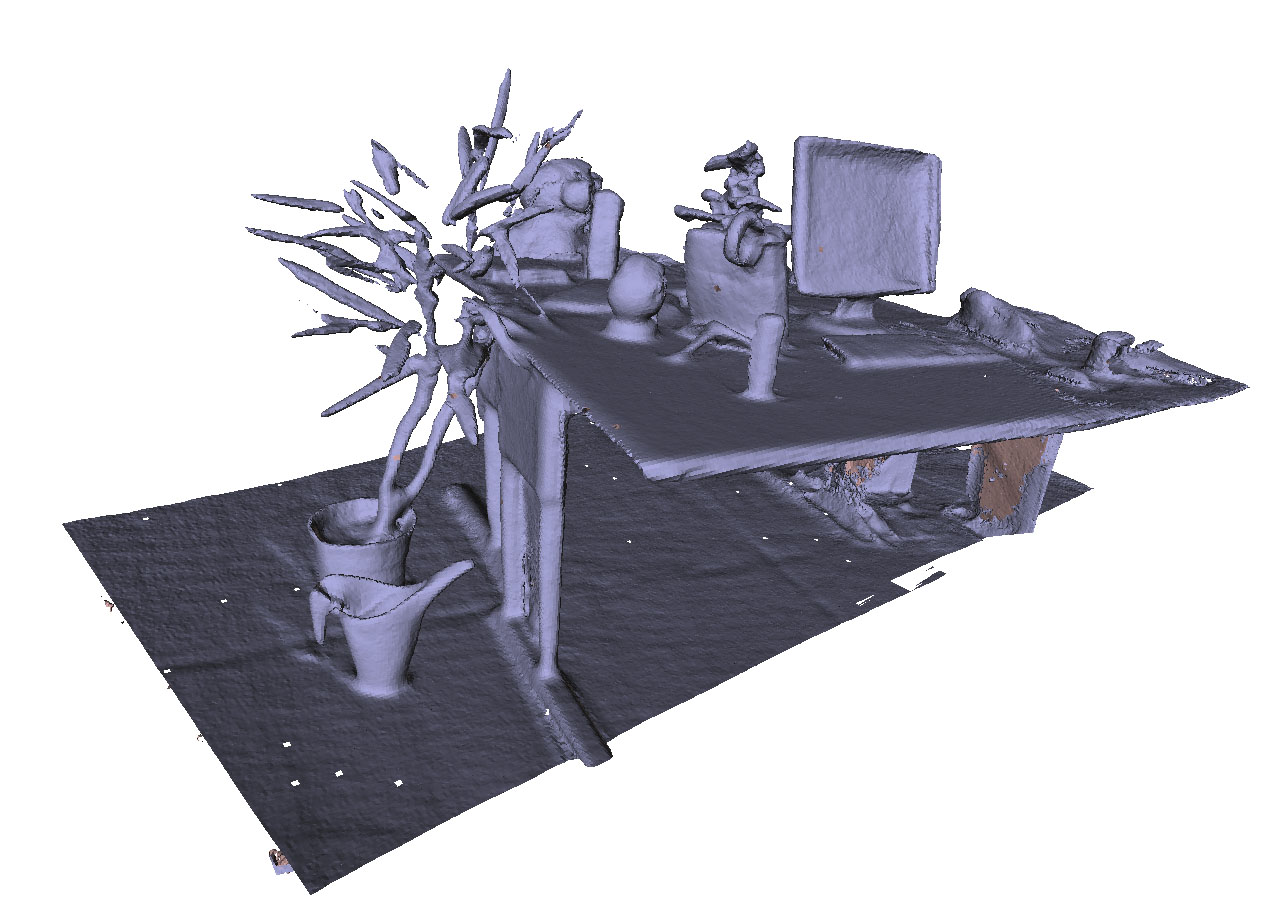}}          
        \subfloat[freiburg1\_room sequence reconstruction.]{ \includegraphics[scale=0.115]{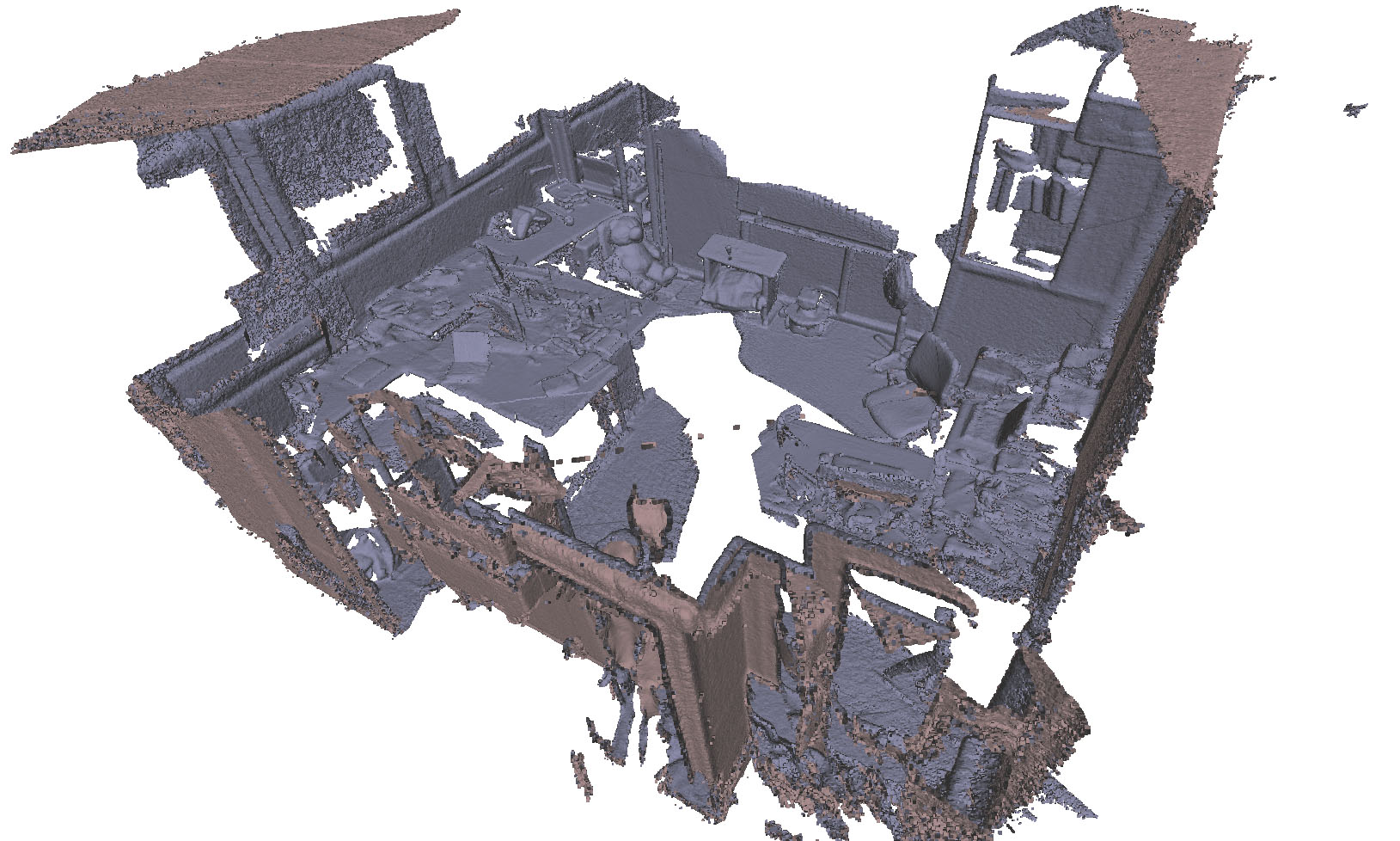}}
               \caption{\small Polygon mesh representation of the RGB-D Benchmark sequences after successfully applying our proposed approach.} 
        \label{fig:meshes}
\end{figure}

After obtaining the final world model, the polygon mesh is extracted by applying the marching cubes algorithm to the voxelized grid representation of the 3D reconstruction \cite{Lorensen87marchingcubes}. The marching cubes algorithm extracts a polygon mesh by subdividing the points cloud or set of 3D points into small cubes (voxels) and marching through each of these cubes to set polygons that represent the isosurface of the points lying within the cube. This results in a smooth surface that approximates the isosurface of the voxelized grid representation, as can be seen in Fig. \ref{fig:meshes}.

%%%%%%%%%%%%%%%%%%%%%%%%%%%%%%%%%%%%%%%%%%%%%%%%%%%%%%%%%%%%%%%%%%%%%%%%%%%%%%%%%%%%%%%%%%%%%%%%%%%%%%%%%%%%%%%%%%%%%%
\section{EXPERIMENTAL RESULTS}
\label{sec:evaluation}
To validate the capabilities of our proposed algorithm, we conduct our evaluation on the RGB-D SLAM Dataset and Benchmark provided by \cite{Benchmark}. This dataset contains RGB-D data and ground-truth data targeted at the evaluation of visual odometry and visual SLAM systems. The data was recorded at the full frame rate of a Microsoft Kinect (30 Hz). The ground-truth trajectory was recorder from motion capture system with eight high-speed tracking cameras  at 100 Hz. We chose to use two specific sequences of the \emph{Handheld SLAM} category: (i) freiburg2\_desk and (ii) freiburg1\_room.  The first sequence is a recording of a typical office scene with two desks, a computer monitor, keyboard, phone, chairs, etc. The second sequence is a recording of a trajectory through a whole office. In both sequences the Kinect is moved around the area and the loop is closed, so we can compare our results with the publicly available trajectories estimated by the RGB-D SLAM system presented by \cite{Endres}.

\subsection{Performance}
The proposed indoor mapping algorithm was tested on a laptop running Ubuntu 12.10 with an Intel Core i7-3720QM processor, 24GB of RAM and an NVIDIA¨ GeForce¨ GTX 670M with 3GB GDDR5 VRAM. With these specifications the original KinectFusion algorithm runs at $\sim$10fps, on the other hand the pure 6D RGB-D odometry method runs as $\sim$5 fps, consequently the runtime of our combined proposed approach is $\sim$3.3 fps. We have identified that the most time consuming processes of the 6D RGB-D odometry approach are the computation of the NARF keypoints on the range image, normal estimation of the depth images and multiple data conversions from PCL to OpenCV format for the BRAND feature computation. Even though our target is not real-time performance but accuracy and map consistency, by optimizing the implementation we predict that the algorithm can reach real-time performance. Furthermore, the current processing time per frame is equal to that of the RGB-D SLAM system proposed by \cite{Endres}.

\subsection{Evaluation}   
\begin{figure}[h]
        \centering
        \subfloat[Kinfu Large Scale presenting drift and failing.]{\includegraphics[scale=0.073]{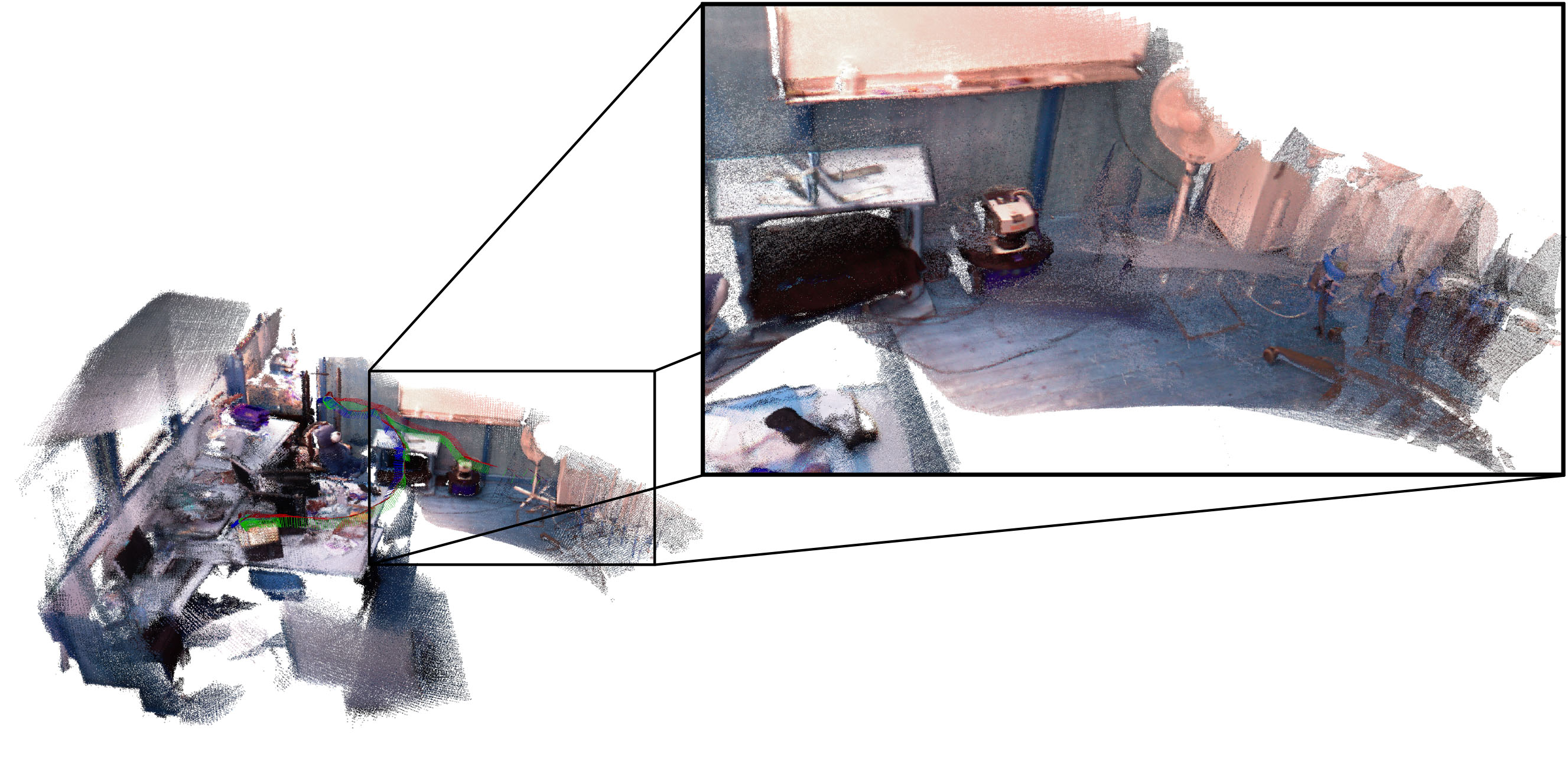}}         
        \subfloat[Consistent mapping with our proposed approach.]{ \includegraphics[scale=0.073]{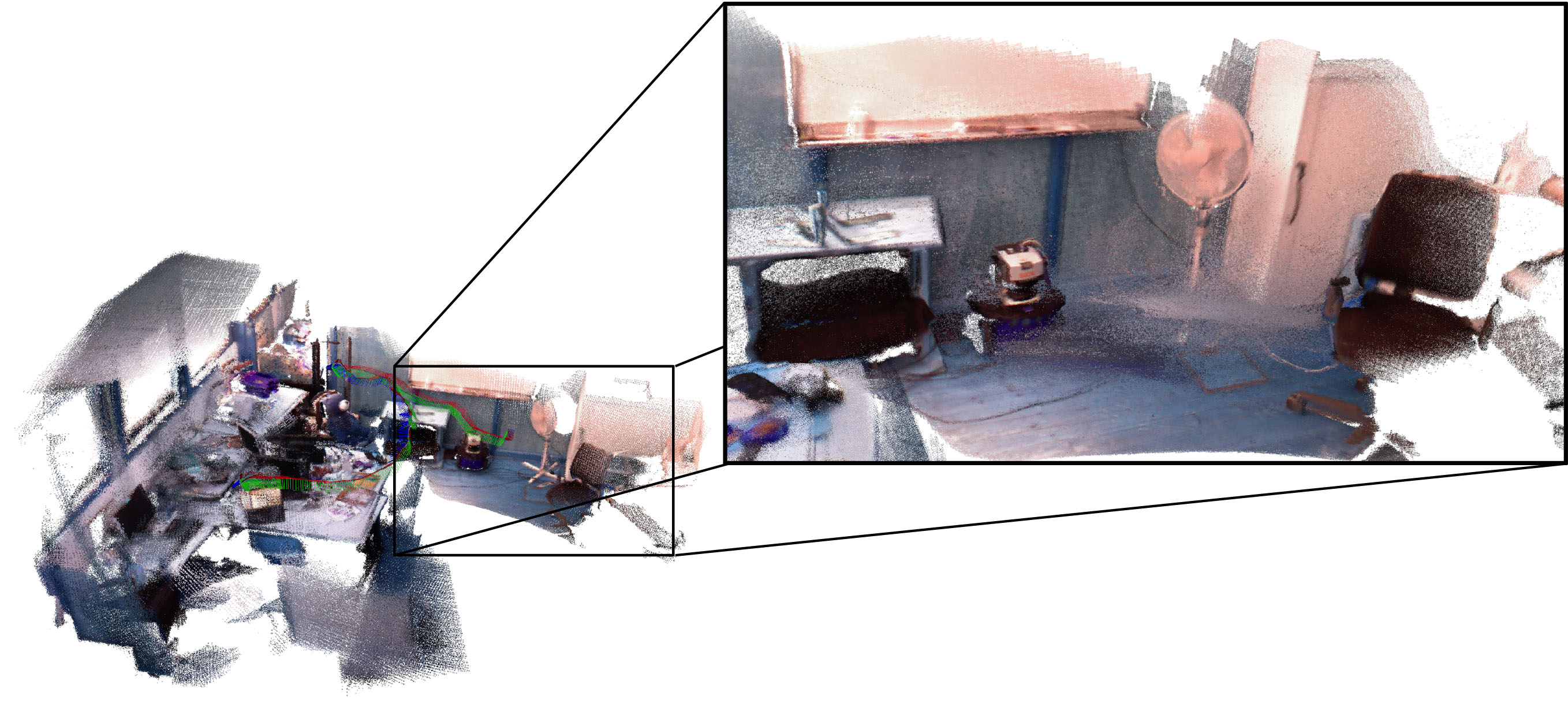}}
        \caption{\small Comparison of pose estimation methods on freiburg1\_room sequence from pose 1-540. Zoomed in section are poses 530-540.}
        \label{fig:kinfu_comparison}
\end{figure}

As mentioned previously, we have evaluated our approach on the freiburg2\_desk and freiburg1\_room sequences. KinFu Large Scale$^3$ performs properly in the freiburg2\_desk sequence and yields to nearly the same accuracy as our approach. However, when applied on the freiburg1\_room sequence, it fails to incrementally estimate the camera poses due to the planar surfaces of the walls and closets. This can be seen in Fig. \ref{fig:kinfu_comparison}. KinFu Large Scale performs quite well up until pose 530, which is when the dense ICP starts computing pose estimates with drift and finally fails in finding an estimate at around camera pose 540. We illustrate this same trajectory with our combined approach. As can be seen, our 6D RGB-D odometry method in combination with KinFu Large Scale, shows no sign of drift and moreover alleviates the failure from using dense ICP based solely on the depth information. 

In order to provide a substantial numerical comparison, we evaluate the quality of the estimated camera trajectories on the dataset sequences using the automatic evaluation tool for RGB-D SLAM results, provided by the RGB-D benchmark. We compare the absolute trajectory error [m] and the relative pose error [m and $^\circ$] of the estimated camera trajectories with the ground truth data collected from a highly accurate motion capture system. In order to have a fair comparison between both methods, we compute the relative pose error for pose pairs with a distance of 1s between them. In Table \ref{tab:results} we present several error metrics (Std - standard deviation, Median, RMS - root mean square, Mean, Max - maximum) based on the absolute and relative trajectory errors. In the freiburg2\_desk sequence our method shows comparable results with the RGB-D SLAM accuracy, we provide some minor improvements in the absolute trajectory error. This behavior was predicted, since this dataset has many salient visual as well as depth features, both algorithms perform well. Furthermore, the trajectory resembles more or less a circular loop with minimum change in elevation and rotation. These characteristics in a trajectory enable the pose graph optimization to be quite straight-forward, instead of distributing the error it is minimized appropriately.

\begin{table}[h]
\caption{\small Comparative results of absolute and relative trajectory errors against ground truth data.}
\label{tab:results}
%\tiny
\centering
\begin{tabular}{ llcccc}
\hline%
\multicolumn{2}{ c } {} & \multicolumn{2}{c}{freiburg2\_desk}  & \multicolumn{2}{c}{freiburg1\_room}\\
\multicolumn {2}{ c }{Sequence} &  RGB-D  &  Our  &  RGB-D  &  Our\\
		&                 &  SLAM &  Approach    	&  SLAM &  Approach\\\hline\hline
		 & Std         & 0.0160 &  0.0509	         & 0.0706 & 0.0833  \\\cline {2 -6}
Absolute    & Median & 0.0946 & {\bf 0.0671}	& 0.0487 & 0.1735   \\\cline{2-6}
Trajectory  & RMS      & 0.0950 & {\bf 0.0949}    & 0.1011 & 0.1972   \\\cline{2-6}
Error [m]     & Mean     & 0.0936 & {\bf 0.0801}   & 0.0723 & 0.1787   \\\cline{2-6}
		  & Max       & 0.1458 & 0.2543   	& 0.4365 & {\bf 0.4164}\\\hline\hline

		& Std 	& 0.0083 	& 0.0171  	& 0.0784 & {\bf 0.0308}\\\cline {2 -6}
Relative 	& Median & 0.0130 	& 0.0145   	& 0.0363 &  0.0461\\\cline {2 -6}
Pose  	& RMS 	& 0.0167 	&  	 0.0258     & 0.0953 & {\bf 0.0595} \\\cline{2-6}	
Error [m] 	& Mean 	& 0.0144	&       0.0194  	& 0.0542 & {\bf 0.0509}  \\\cline{2-6}
		& Max 	& 0.0926 	& 0.1208   	& 0.5580 & {\bf 0.1615}   \\\hline\hline
		
			& Std 	& 0.3294 &  0.7983  	& 2.0767 	& {\bf 1.3037} \\\cline {2 -6}
Relative 		& Median 	& 0.0089 & 0.0123  	& 0.0337 	&  0.0446 \\\cline {2 -6}
Pose  		& RMS 	& 0.6607 &  1.2329  	& 3.1566  & {\bf 2.9982}\\\cline{2-6}
Error [$^\circ$] 	& Mean 	& 0.5727 &  0.9395     & 2.3772 	&  2.6999 \\\cline{2-6}
			& Max 	& 3.0105 &  6.1661  	& 13.987 	& {\bf 8.6952} \\\hline
\end{tabular}
\end{table}

Nonetheless, for more difficult scenes and trajectories (like the freiburg1\_room sequence) our method not only improves the behavior of KinFu Large Scale but it also outperforms RGB-D SLAM.  We provide a maximum relative pose translational error of 0.16m and rotational of $8.6^\circ$ as opposed to 0.55m and $13.98^\circ$ from RGB-D SLAM. Clearly, for extended spaces and rooms where there is a lack of descriptive surfaces, our approach gives the most consistent trajectories and maps, as can be seen in Fig.\ref{fig:meshes}. Let's recall that these results are after loop closure and graph optimization regarding the RGB-D SLAM approach. This means that graph pose optimization was not able to completely minimize the propagated errors from visual odometry, instead it distributed the errors throughout the trajectory. The differences between the estimated trajectories of each method and the ground truth can be found in the appendix. 

%%%%%%%%%%%%%%%%%%%%%%%%%%%%%%%%%%%%%%%%%%%%%%%%%%%%%%%%%%%%%%%%%%%%%%%%%%%%%%%%
\section{CONCLUSION}
\label{sec:conclusion}
In this work, we presented our first efforts into creating a consistent reconstruction of indoor spaces by augmenting the highly accurate KinectFusion with a novel 6D RGB-D odometry algorithm. We tested our combined approach on a notable RGB-D Benchmark dataset and present highly encouraging results. We provide nearly half of the absolute and relative pose errors from a state-of-the-art RGB-D SLAM system. An important contribution of this paper is that we demonstrate that the fusion of color and depth information yields to robust frame-to-frame alignment as opposed to solely using RGB (as in typical visual odometry approaches) or depth (as in the KinectFusion algorithm), which is the most important component of a 3D mapping application. Furthermore, the use of a consumer device such as the Microsoft Kinect opens up the possibility of open source personal 3D mapping, i.e. reconstructing your own environment with freely available software and sharing the reconstructed environments for many possible applications such as creating interactive virtual worlds. 

% Bibliography
\bibliographystyle{ACM-Reference-Format-Journals}
\bibliography{literature}

%%% -*-BibTeX-*-
%%% Do NOT edit. File created by BibTeX with style
%%% ACM-Reference-Format-Journals [18-Jan-2012].

\begin{thebibliography}{00}

%%% ====================================================================
%%% NOTE TO THE USER: you can override these defaults by providing
%%% customized versions of any of these macros before the \bibliography
%%% command.  Each of them MUST provide its own final punctuation,
%%% except for \shownote{}, \showDOI{}, and \showURL{}.  The latter two
%%% do not use final punctuation, in order to avoid confusing it with
%%% the Web address.
%%%
%%% To suppress output of a particular field, define its macro to expand
%%% to an empty string, or better, \unskip, like this:
%%%
%%% \newcommand{\showDOI}[1]{\unskip}   % LaTeX syntax
%%%
%%% \def \showDOI #1{\unskip}           % plain TeX syntax
%%%
%%% ====================================================================

\ifx \showCODEN    \undefined \def \showCODEN     #1{\unskip}     \fi
\ifx \showDOI      \undefined \def \showDOI       #1{{\tt DOI:}\penalty0{#1}\ }
  \fi
\ifx \showISBNx    \undefined \def \showISBNx     #1{\unskip}     \fi
\ifx \showISBNxiii \undefined \def \showISBNxiii  #1{\unskip}     \fi
\ifx \showISSN     \undefined \def \showISSN      #1{\unskip}     \fi
\ifx \showLCCN     \undefined \def \showLCCN      #1{\unskip}     \fi
\ifx \shownote     \undefined \def \shownote      #1{#1}          \fi
\ifx \showarticletitle \undefined \def \showarticletitle #1{#1}   \fi
\ifx \showURL      \undefined \def \showURL       #1{#1}          \fi

\bibitem[\protect\citeauthoryear{Agrawal, Konolige, and Blas}{Agrawal
  et~al\mbox{.}}{2008}]%
        {STAR}
{Motilal Agrawal}, {Kurt Konolige}, {and} {Morten~Rufus Blas}. 2008.
\newblock \showarticletitle{CenSurE: Center surround extremas for realtime
  feature detection and matching}. In {\em Proceedings of the European
  Conference on Computer Vision}. 102--115.
\newblock


\bibitem[\protect\citeauthoryear{Ali, Shafait, Giannakidou, Vakali, Figueroa,
  Varvadoukas, and Mavridis}{Ali et~al\mbox{.}}{2013}]%
        {RAS-Ali}
{Haider Ali}, {Faisal Shafait}, {Eirini Giannakidou}, {Athena Vakali}, {Nadia
  Figueroa}, {Theodoros Varvadoukas}, {and} {Nikolaos Mavridis}. 2013.
\newblock \showarticletitle{Contextual object category recognition in {RGB-D}
  scene labeling}.
\newblock {\em To appear in Robotics and Autonomous Systems\/} (2013).
\newblock


\bibitem[\protect\citeauthoryear{Bay, Ess, Tuytelaars, and van Gool}{Bay
  et~al\mbox{.}}{2008}]%
        {Bay-CVIU08}
{Herbet Bay}, {Andreas Ess}, {Tinne Tuytelaars}, {and} {Luc van Gool}. 2008.
\newblock \showarticletitle{{SURF}: Speeded up robust features}.
\newblock {\em Computer Vision and Image Understanding\/} {110}, 3 (2008),
  346--359.
\newblock


\bibitem[\protect\citeauthoryear{Besl and McKay}{Besl and McKay}{1992}]%
        {Besl92}
{Paul~J. Besl} {and} {Neil~D. McKay}. 1992.
\newblock \showarticletitle{A method for registration of 3{D} shapes}.
\newblock {\em IEEE Transactions on Pattern Analysis and Machine
  Intelligence\/} {14}, 2 (1992), 239--256.
\newblock


\bibitem[\protect\citeauthoryear{Bradski}{Bradski}{2000}]%
        {opencv_library}
{Gary Bradski}. 2000.
\newblock \showarticletitle{{The OpenCV library}}.
\newblock {\em Dr. Dobb's Journal of Software Tools\/} (2000).
\newblock


\bibitem[\protect\citeauthoryear{Chen and Medioni}{Chen and Medioni}{1991}]%
        {Chen91}
{Yang Chen} {and} {G\'erard Medioni}. 1991.
\newblock \showarticletitle{Object modeling by registration of multiple range
  images}. In {\em Proceedings of IEEE International Conference on Robotics and
  Automation}. 2724 --2729.
\newblock


\bibitem[\protect\citeauthoryear{Curless and Levoy}{Curless and Levoy}{1996}]%
        {SDF}
{Brian Curless} {and} {Marc Levoy}. 1996.
\newblock \showarticletitle{A volumetric method for building complex models
  from range images}. In {\em Proceedings of the 23rd Annual Conference on
  Computer Graphics and Interactive Techniques}. 303--312.
\newblock


\bibitem[\protect\citeauthoryear{Endres, Hess, Engelhard, Sturm, Cremers, and
  Burgard}{Endres et~al\mbox{.}}{2012}]%
        {Endres}
{Felix Endres}, {J\"urgen Hess}, {Nikolas Engelhard}, {J\"urgen Sturm}, {Daniel
  Cremers}, {and} {Wolfram Burgard}. 2012.
\newblock \showarticletitle{An evaluation of the RGB-D SLAM system}. In {\em
  Proceedings of the IEEE International Conference on Robotics and Automation}.
  1691--1696.
\newblock


\bibitem[\protect\citeauthoryear{Figueroa, Ali, and Schmidt}{Figueroa
  et~al\mbox{.}}{2012}]%
        {Figueroa-CRV2012}
{Nadia Figueroa}, {Haider Ali}, {and} {Florian Schmidt}. 2012.
\newblock \showarticletitle{3D registration for verification of humanoid
  Justin's upper body kinematics}. In {\em Proceedings of the Ninth Conference
  on Computer and Robot Vision}. 276--283.
\newblock


\bibitem[\protect\citeauthoryear{Lorensen and Cline}{Lorensen and
  Cline}{1987}]%
        {Lorensen87marchingcubes}
{William~E. Lorensen} {and} {Harvey~E. Cline}. 1987.
\newblock \showarticletitle{Marching cubes: A high resolution 3{D} surface
  construction algorithm}.
\newblock {\em Computer Graphics\/} {21}, 4 (1987), 163--169.
\newblock


\bibitem[\protect\citeauthoryear{Lowe}{Lowe}{2004}]%
        {Lowe-IJCV04}
{David~G. Lowe}. 2004.
\newblock \showarticletitle{Distinctive image features from scale-invariant
  keypoints}.
\newblock {\em International Journal of Computer Vision\/} {60}, 2 (2004),
  91--110.
\newblock
\showISSN{0920-5691}


\bibitem[\protect\citeauthoryear{Nascimento, Oliveira, Campos, Vieira, and
  Schwartz}{Nascimento et~al\mbox{.}}{2012}]%
        {Brand}
{Erickson~R. Nascimento}, {Gabriel~L. Oliveira}, {Mario F.~M. Campos},
  {Ant\^onio~W. Vieira}, {and} {Williamson~Robson Schwartz}. 2012.
\newblock \showarticletitle{BRAND: A robust appearance and depth descriptor for
  {RGB-D} images}. In {\em Proceedings of the IEEE/RSJ International Conference
  on Intelligent Robots and Systems}. 1720--1726.
\newblock
\showISSN{2153-0858}


\bibitem[\protect\citeauthoryear{Newcombe, Izadi, Hilliges, Molyneaux, Kim,
  Davison, Kohli, Shotton, Hodges, and Fitzgibbon}{Newcombe
  et~al\mbox{.}}{2011}]%
        {KinectFusion}
{Richard~A. Newcombe}, {Shahram Izadi}, {Otmar Hilliges}, {David Molyneaux},
  {David Kim}, {Andrew~J. Davison}, {Pushmeet Kohli}, {Jamie Shotton}, {Steve
  Hodges}, {and} {Andrew~W. Fitzgibbon}. 2011.
\newblock \showarticletitle{KinectFusion: Real-time dense surface mapping and
  tracking}. In {\em Proceedings of the 10th IEEE International Symposium on
  Mixed and Augmented Reality}. 127--136.
\newblock


\bibitem[\protect\citeauthoryear{Scaramuzza and Fraundorfer}{Scaramuzza and
  Fraundorfer}{2011}]%
        {VO}
{Davide Scaramuzza} {and} {Friedrich Fraundorfer}. 2011.
\newblock \showarticletitle{Visual odometry [Tutorial]}.
\newblock {\em IEEE Robotics Automation Magazine\/} {18}, 4 (2011), 80--92.
\newblock
\showISSN{1070-9932}


\bibitem[\protect\citeauthoryear{Sturm, Engelhard, Endres, Burgard, and
  Cremers}{Sturm et~al\mbox{.}}{2012}]%
        {Benchmark}
{J\"urgen Sturm}, {Nikolas Engelhard}, {Felix Endres}, {Wolfram Burgard}, {and}
  {Daniel Cremers}. 2012.
\newblock \showarticletitle{A benchmark for the evaluation of RGB-D SLAM
  systems}. In {\em Proceedings of the IEEE/RSJ International Conference on
  Intelligent Robots and Systems}. 573--580.
\newblock


\bibitem[\protect\citeauthoryear{Umeyama}{Umeyama}{1991}]%
        {Umeyama}
{Shinji Umeyama}. 1991.
\newblock \showarticletitle{Least-squares estimation of transformation
  parameters between two point patterns}.
\newblock {\em IEEE Transactions on Pattern Analysis and Machine
  Intelligence\/} {13}, 4 (1991), 376--380.
\newblock
\showISSN{0162-8828}


\bibitem[\protect\citeauthoryear{Whelan, Johannsson, Kaess, Leonard, and
  McDonald}{Whelan et~al\mbox{.}}{2012}]%
        {Kintinuous2}
{Thomas Whelan}, {Hordur Johannsson}, {Michael Kaess}, {John~J. Leonard}, {and}
  {John McDonald}. 2012.
\newblock {\em Robust tracking for real-time dense {RGB-D} mapping with
  {K}intinuous}.
\newblock {T}echnical {R}eport MIT-CSAIL-TR-2012-031. Computer Science and
  Artificial Intelligence Laboratory, MIT.
\newblock


\bibitem[\protect\citeauthoryear{Yang}{Yang}{2012}]%
        {eccv-12-qingxiong-yang}
{Qing~Xiong Yang}. 2012.
\newblock \showarticletitle{Recursive bilateral filtering}. In {\em Proceedings
  of the 12th European Conference on Computer Vision}. 399--413.
\newblock


\bibitem[\protect\citeauthoryear{Zhang}{Zhang}{1994}]%
        {Zhengyou94}
{Zhengyou Zhang}. 1994.
\newblock \showarticletitle{Iterative point matching for registration of
  free-form curves and surfaces}.
\newblock {\em {International Journal of Computer Vision}\/} {13}, 2 (1994),
  119--152.
\newblock


\end{thebibliography}
                             % Sample .bib file with references that match those in
                             % the 'Specifications Document (V1.5)' as well containing
                             % 'legacy' bibs and bibs with 'alternate codings'.
                             % Gerry Murray - March 2012

% History dates
\received{December 16, 2013}{}{March 22, 2014}

% Electronic Appendix
\elecappendix

\medskip
\section{Umeyama Rigid Motion Derivation}
This method demeans the 3D point sets as 
\begin{equation}
\begin{array}{rcl} 
p'_{k-1,i} & = & p_{k-1,i} - \overline{p}_{k-1} \\
p'_{k,i}    & = & p_{k,i} - \overline{p}_{k}  \\  
\forall_{i}: p'_{k,i}  & = & Rp'_{k-1,i}  \\
P'_{k-1} & = & RP'_{k},
\end{array}
\end{equation}
and computes the rotational component with the Singular Value Decomposition of the correlation matrix and the translation component based on the mean positions as follows
\begin{equation}
\begin{array}{rcl} 
svd(P_{k-1}',P_{k}')^{T} & = & UDV^{T} \\ 
R & = & VU^T \\ 
t  & = & \overline{p}_{k-1}  - R \overline{p}_{k}.
\end{array}
\end{equation}

\section{Rigid Motion Estimation Algorithm}
\begin{algorithm}[h]
\renewcommand{\algorithmicrequire}{\textbf{Input:}}
\renewcommand{\algorithmicensure}{\textbf{Output:}}
\caption{Iterative Rigid Motion Estimation}
\label{alg:rigidmotion}
\begin{algorithmic} 
\REQUIRE Corresponding matches $MM_{k-1}^{k}$, source point cloud $P_{k-1}$, target point cloud $P_{k}$, maximum number of hypotheses iterations $N$, number sample correspondences $s$ and error threshold $e_t$
\ENSURE A rigid transformation ${T^{k}_{k-1}}_{rgbd}$ that aligns $P_{k-1} \rightarrow P_{k}$
\FOR{i in N}
		\STATE$ mm_i \leftarrow  findRandomCombinations(MM_{k-1}^{k},s)$
		\STATE$ T_i \leftarrow  computeUmeyamaRigidMotion(P_{k-1},P_{k} ,mm_i)$
		\STATE $e_i \leftarrow  computeTransformationError(P_{k-1},P_{k} T_i)$
\ENDFOR
\STATE ${T^{k}_{k-1}}_{rgbd} \leftarrow  findMinimumTransformationError(e_i \in e$ and $T_i \in T$ for $i=1..N)$
\end{algorithmic}
\end{algorithm}

\section{Evaluation}
For each sequence we provide two plots: (i) the trajectory difference and (ii) the relative translational error. These trajectory differences represent the pose pair errors between the estimated and ground truth trajectories and is illustrated with a red vector connecting the pose pairs. 
\begin{figure}[h]
\centering
        \subfloat[Trajectory difference with our approach.]{\includegraphics[scale=0.32]{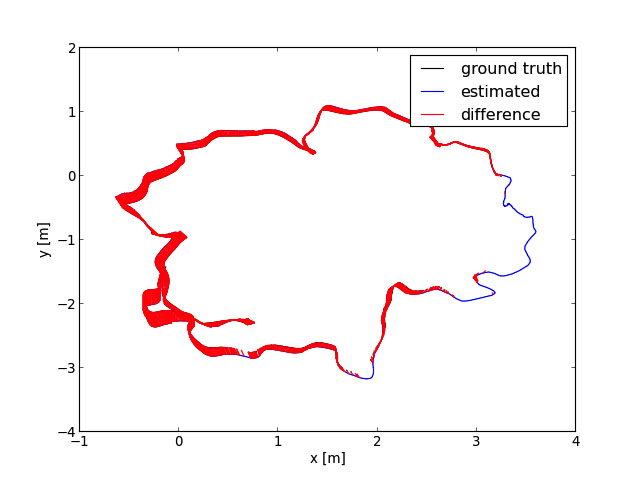}}       
        \subfloat[Translational errors of our approach.]{ \includegraphics[scale=0.32]{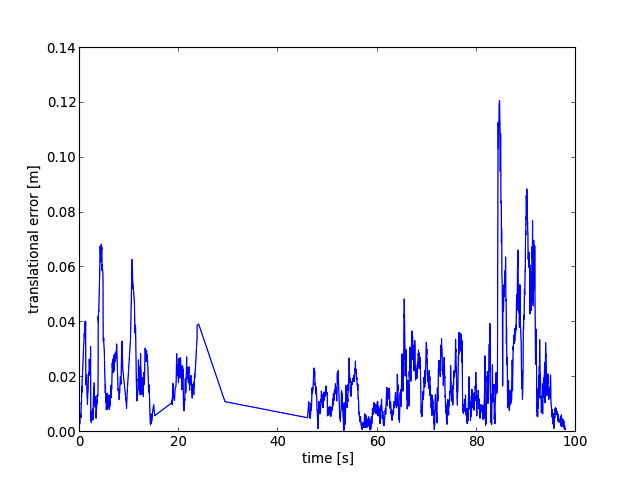}}
        
       \subfloat[Trajectory difference with RGB-D SLAM.]{\includegraphics[scale=0.32]{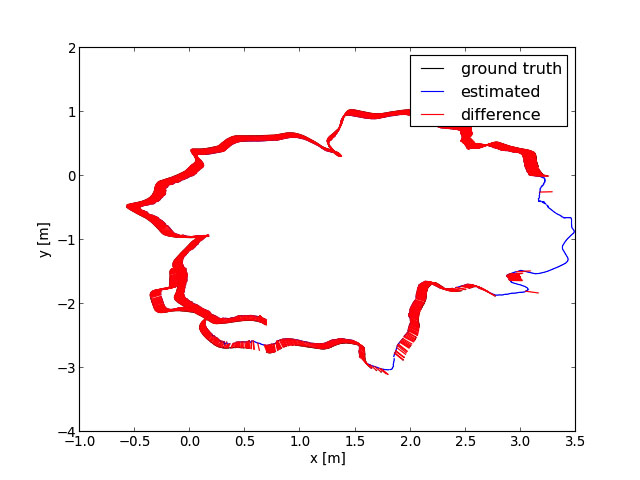}}       
        \subfloat[Translational errors of RGB-D SLAM.]{ \includegraphics[scale=0.32]{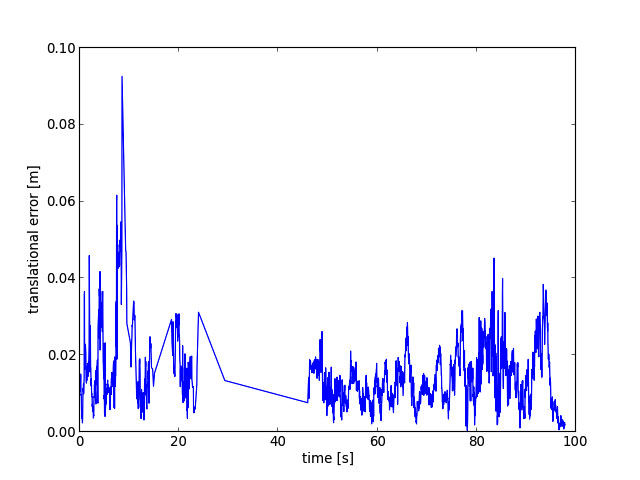}}      
        \caption{Estimated trajectories (a,c) of the freiburg2\_desk sequence compared to the ground truth and relative translational errors (b,d) for RGB-D SLAM (top row) and our approach (bottom row).}\label{fig:desk_trajectories}
\end{figure}

\begin{figure}[h]
        \centering
        \subfloat[Trajectory difference with our approach.]{\includegraphics[scale=0.32]{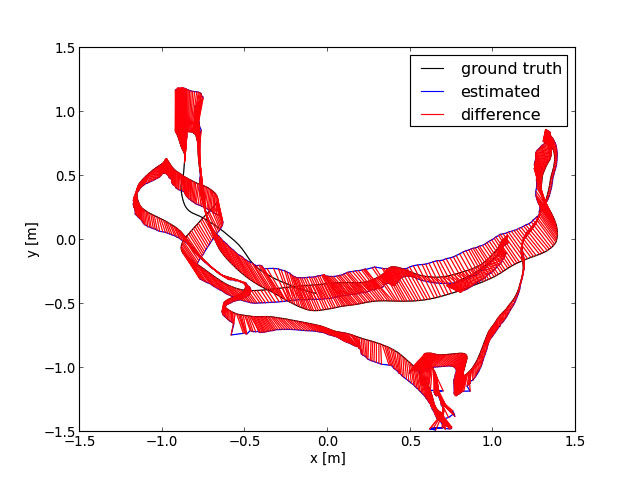}}       
        \subfloat[Translational errors of our approach.]{ \includegraphics[scale=0.32]{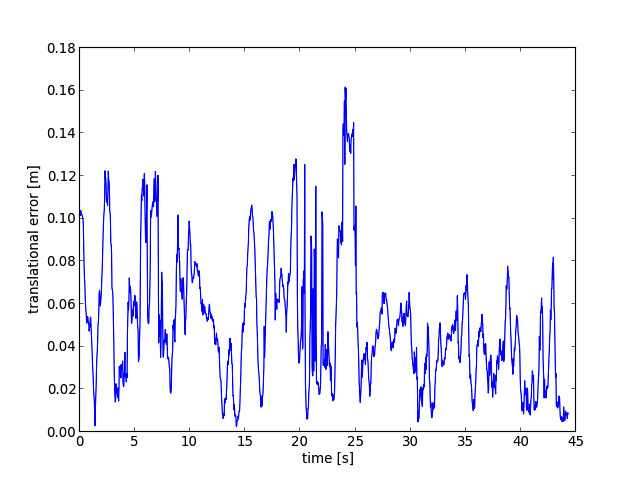}}
        
       \subfloat[Trajectory difference with RGB-D SLAM.]{\includegraphics[scale=0.32]{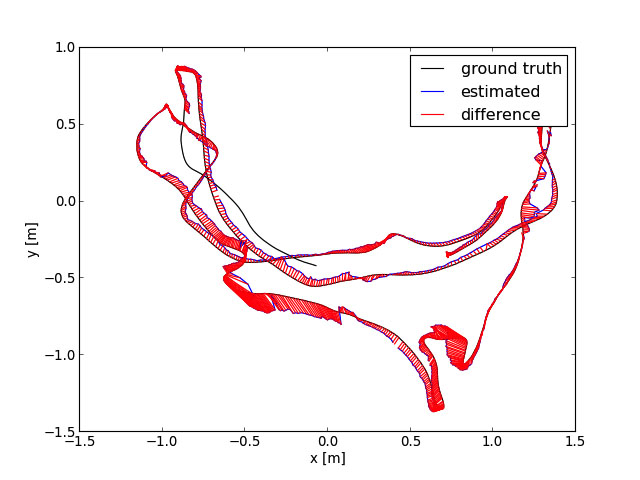}}       
        \subfloat[Translational errors of RGB-D SLAM.]{ \includegraphics[scale=0.32]{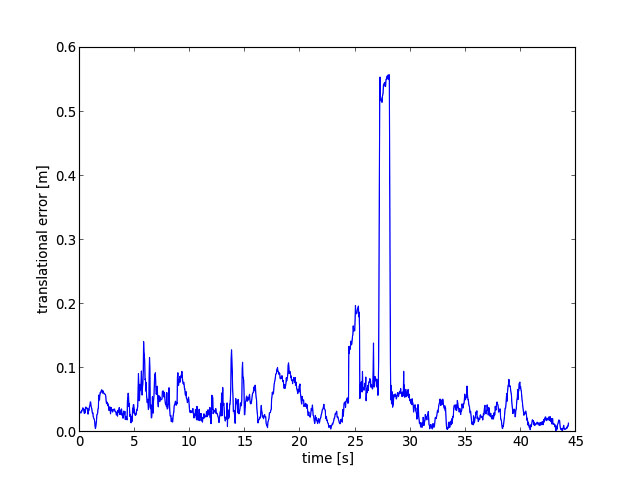}}      
        \caption{Estimated trajectories of the freiburg1\_room sequence compared to the ground truth and relative translational errors (b,d) for RGB-D SLAM (top row) and our approach (bottom row).}\label{fig:room_trajectories}
\end{figure}

\end{document}